# Stemmer for Serbian language

Nikola Milošević

http://www.inspiratron.org

## Abstract

In linguistic morphology and information retrieval, stemming is the process for reducing inflected (or sometimes derived) words to their stem, base or root form—generally a written word form. In this work is presented suffix-stripping stemmer for Serbian language, one of the highly inflectional languages.

## 1. Introduction

When performing many of the natural language processing tasks it is needed that all forms of a word with same meaning has same form. This process of fitting words to usable forms is called normalization. There are several techniques to perform normalization such as normalization using equivalence classes, case-folding, stemming and lemmatization.

*Stemming* usually refers to a crude heuristic process that chops off the ends of words in the hope of achieving this goal correctly most of the time, and often includes the removal of derivational affixes.

*Lemmatization* usually refers to doing things properly with the use of a vocabulary and morphological analysis of words, normally aiming to remove inflectional endings only and to return the base or dictionary form of a word, which is known as the *lemma*[1].

Stemming and lemmatization have wide use in many natural language processing tasks. They are used retrieve semantic and meaning from different word's forms. Most of natural language processing tasks are based on retrieving the meaning of processed text, but are not interested in other information that gives different word's forms. Task with applied stemming and lemmatization are information retrieval, machine translation, web or document search, text classification and sentiment analysis as a special case of text classification.

In this work is introduced stemmer for Serbian language, which will be used for sentiment analysis. To build successful lemmatization tool it is required dictionary with all forms of all words in particular language. Obviously, it requires greater development time and big database. Since there is no significant performance difference between stemming and lemmatization [2], it is much easier to build stemmer. Sentiment analysis has some special requirements for stemming, so this stemmer has to be modified for use in other applications such as information retrieval. For example stemmer that will be used in sentiment analysis will not remove prefixes, since they are important part of sentiment in Serbian language (i.e. superlative prefixes).

## 2. History

The first ever written stemming algorithm was written by Julie Beth Lovins in 1968.[3]. Most popular stemmer for English was written by Martin Porter in 1980. This stemmer became de-facto standard for English stemming and Martin Porter was rewarded with Tony Kent Strix award for his work on stemming and information retrieval in year 2000. Until year 2000 almost all works related to stemming was about stemming for English, since 60% of World Wide Web is on English language. But after year 2000, there was emerging need to build natural language processing application also for other languages. In year 2000, Martin Porter published his work on snowball framework for creating stemming algorithms, and published stemming algorithms for several languages. Rules for stemming of Serbian language are still not precisely defined.

## 3. Serbian language

Serbian language is a form of South Slavic language (Indo-European), spoken mainly by Serbs. All South Slavic languages derive their forms from Old Church Slavonic with Serbian emerging from Old Serbian (Serbian-Slavonic), which has a literary history from 10th century.

Serbian language represents highly inflectional language. Because of this characteristic stemmer for Serbian language will have much more rules then stemmers for less inflectional languages like English. In Porter stemmer total number of rules is 63, but if we represent it in plain suffix format we got about 120 rules [4]. Stemmer for Serbian language will have between 3 and 4 times more suffix stripping rules then stemmer for English because of its higher inflection.

## 4. Related work

As previously mentioned Lovins and Porter's stemmers are the first stemmers ever made. In last 10 years there were created stemmers for many other languages like German, French, Spanish, Portuguese, Italian, Romanian, Dutch, Swedish, Norwegian, Danish, Russian, Finish, Czech, Irish, Hungarian, Turkish, Armenian, Basque, Catalan using snowball framework [6]. In this list there are only two Slavic languages - Russian and Czech. Search for stemmers of south Slavic languages gave just few results. There is stemmer for Slovenian language [7]. Also there is stemmer for Croatian, but with little explanation found on SVN repository

(http://svn.rot13.org/index.cgi/stem-hr). For Serbian language there is work called "A suffix subsumption -based approach to building stemmers and lemmatizers for highly inflectional languages with sparse resource" by Vlado Kešelj and Danko Šipka [4]. In the paper authors stated that their stemmer has accuracy of 79%. Investigating it was realized that accuracy of this stemmer can be fine tuned manually, and that number of rules can be much reduced from more than 1000 stemming rules in Optimal Suffix stemmer to less than 300 stemming rules. This optimization will eventually improve speed and accuracy of stemmer. Stemmer developed by Vlado Kešelj and Danko Šipka was used as base for our stemmer.

## 5. Stemming application and algorithm

Stemming application is designed as web application. It uses PHP script for backend and AJAX interaction with backend side. Application has two text boxes, one for text input, and the other will be filled as output of stemmer. Stemmer will perform his work pressing on button "Stem".

When button "Stem" is pressed, first action that will be taken is transforming of special Serbian characters such as č, ć, š, ž, đ into cx, cy, sx, zx, and dx. When client receives response, he will also using JavaScript transform special characters back to normal, readable form. This will simplify text manipulation and transferring text between client and server. After this transformation client creates post request with text as his post parameter queue.

PHP script will do several transformations to the text. This will simplify text processing. Script will change all upper case letters to lower case (case folding). Also it will add space character before and after all punctuation marks. This will make punctuation marks separate tokens. After this, tokenization is performed.

There exist two approaches for stemming. One is algorithmic, while the other is dictionary based. To obtain perfect stems for Serbian language it is has to be applied hybrid approach. Most of words can be stemmed using suffix stripping algorithm. While some irregular verbs and some other types of words with irregular flections have to be stemmed using dictionary. Algorithm that will be used will look at dictionary if the word exists in it, if it exist it will change it with basic form, otherwise it will look in suffix rule list and strip or change suffix.

Dictionary of irregular words is matching whole words and changing it with others.

Server side PHP script will use regular expressions to match suffixes and strip or change them. Stems should not have less than 2 characters, since they will become unusable, so algorithm is searching for suffixes after second character. This rule has to include special characters in Serbian language, since they are coded to two characters. If special character occurs it will look for suffixes after third character.

Suffix rules are organized as key-value dictionary. Full suffix is a key, and suffix substitution is value. Most of values are empty strings, but since some suffixes need to be changed with some string, key-value dictionary data structure appeared as best solution.

## 6. Development of stemming rules for Serbian language

Development of stemming rules for Serbian language was started with 1000 rules from stemmer developed by Vlado Kešelj and Danko Šipka [4] adjusted for PHP. Since these stemming rules was created by machine learning algorithm it created some erroneous rules. It has also too many rules. Some of erroneous rules included rules that contained x or y character at it beginning. These rules should contain full special character, but machine learning algorithm was not aware about coding of Serbian special characters. There was too many rules, that made in some cases that stemmer will use wrong rule.

First step building stemming rules was to clear the list of rules from erroneous and unneeded rules. Doing this we came to list of about 180 rules. Clearing of unneeded rules were done manually. Rules starting with x or y were deleted, but also rules longer than 5 characters. There were some rules identified as badly formed, because they will strip more than suffix, or that they will strip some part of previously defined stem for that word. It was identified that different forms of same word has different stems. Since stemmer should perform same stems for all forms of same word, this problem had to be fixed by deleting bad stemming rules and adding new good rules. Since it was observed that stemmer cannot stem correctly some forms, it was needed to add some new rules. These newly added rules were not discovered by machine learning algorithm described by Kešelj and Šipka.

Two approaches were used developing new rules for stemmer. First one was consulting grammar of Serbian language about word forms and transformations. This approach improved quality of stems quite well, but it was observed that some of words still are not correctly stemmed. The other approach was to manually identify words that cannot be correctly stemmed and try to create rule for that word form.

Since imperfect tense is no longer used in modern Serbian language and because it's forms are much different from the root of verb, it is not supported in this stemmer.

Dictionary of words was considered as solution only when there was no other acceptable solution for some words and transformations. Since irregular verbs like "biti", "hteti" and "jesam", "moći" has different forms of base in each tense or person, they had to be stemmed using dictionary. There were also other verbs on which dictionary rules were applied like "naći", "doći", "ići", "otići", "stići", "vući", and some more. Also this rules was also applied as suffix changing rules, since there are lot of verbs that is build adding prefix on this verbs like "pronaći", "prevući", "dovući" etc.

# 7. Evaluation

Stemmer for Serbian language is evaluated using 522 word text taken from newspaper "Politika". Two methods of evaluation were performed. In first one, text from newspaper were manually stemmed and compared to stemmer output. Person who manually stemmed the text knew the rules that stemmer was using, but also was taking care about output to be logical and not under stemmed. Second method involves stemming, and then manually reading text. A stem is judged to be correct if the original meaning can be clearly predicted from the stem (no over-stemming), and it seems that the stem covers all morphological variations of the lemma (no under-stemming).

Both evaluation criteria showed very good results. In case of manual stemming, there is a problem that person who stemmed knew the stemming rules, but tried to stay neutral. In this case there were 37 errors. This means that accuracy of stemmer is 92%. We can assume that there will be some additional errors that are not covered, since test text is quite small. But still results are satisfying since they are around 90%. Problematic words that created stems that did not covered all flections were words that contain some voice change. Most problematic was words that contained voice change called "nepostojano a", since rules cannot be created to cover both words that has this voice change and also words that do not. In the other case, where it was tried to retrieve original meaning of word we had similar results. About 90% of words are stemmed in that way that original meaning can be retrieved and stem seemed to cover all morphological variations. Problem was small words, 3-4 characters, which after stemming had only 2. These stems were covering morphological forms, but original meaning was hard to retrieve. But since many of those words can be considered as stop words in sense of information retrieval related tasks, this should not represent big problem.

# 8. Conclusion and further work

Stemmer that is described in this work should be most efficient stemmer for Serbian language so far. Still there is much space for improvement and fine tuning. Still there are some forms of words that are not stemmed correctly and that should be part of future improvement. Since most of words are stemmed correctly, stemmer will be of great help in natural language processing tasks for Serbian language. For Serbian language there are not many build application of natural language processing, so there is hope that this work will be of great help in further work in this field.

This stemmer is built as preprocessing tool for sentiment analysis application. In this sense there might be needed some adjustments for other natural language processing applications such as information retrieval, machine translation or search.

# Appendrix: Stemmer code and rules

| Sending request from stemming form page |
|---|

```javascript
<script language="JavaScript" type="text/javascript"  charset=utf-8">
function stemrequest(str){
if (str.length==0)
    {
        document.getElementById("stem").value="";
        return;
    }
 if (window.XMLHttpRequest)
    {// code for IE7+, Firefox, Chrome, Opera, Safari
         xmlhttp=new XMLHttpRequest();
    }
 else
    {// code for IE6, IE5
        xmlhttp=new ActiveXObject("Microsoft.XMLHTTP");
    }
 xmlhttp.onreadystatechange=function()
    {
        if (xmlhttp.readyState==4 && xmlhttp.status==200)
        {
         rstr =xmlhttp.responseText;
         rstr =rstr.replace(new RegExp("cx","gm"),"č");
         rstr =rstr.replace(new RegExp("cy","gm"),"ć");
         rstr =rstr.replace(new RegExp("zx","gm"),"ž");
         rstr =rstr.replace(new RegExp("sx","gm"),"š");
         rstr =rstr.replace(new RegExp("dx","gm"),"đ");
         document.getElementById("stemed").value=rstr;
        }
    }
xmlhttp.open("POST","SerbTokenizer.php",true);
str=str.replace(new RegExp("č","gm"),"cx");
str=str.replace(new RegExp("ć","gm"),"cy");
str=str.replace(new RegExp("ž","gm"),"zx");
str=str.replace(new RegExp("š","gm"),"sx");
str=str.replace(new RegExp("đ","gm"),"dx");
str=str.replace(new RegExp("Č","gm"),"cx");
str=str.replace(new RegExp("Ć","gm"),"cy");
str=str.replace(new RegExp("Ž","gm"),"zx");
str=str.replace(new RegExp("Š","gm"),"sx");
str=str.replace(new RegExp("Đ","gm"),"dx");
var parameters="queue="+encodeURIComponent(str);
xmlhttp.setRequestHeader("Content-type", "application/x-www-form-urlencoded");
xmlhttp.setRequestHeader("Content-Length", parameters.length);
xmlhttp.setRequestHeader("Connection", "close");
 xmlhttp.send(parameters);
 }
</script>
```

| SerbStemmer.php |
|---|
| ```php
<?php
//Author: Nikola Milosevic, http://www.inspiratron.org
// This work is done as part of master thesis of Nikola Milosevic on
//University of Belgrade, School of Electrical Engineering
//This program is free software: you can redistribute it and/or modify it
//under the terms of the GNU General Public License as published by the
//Free Software Foundation, either version 3 of the License or (at your
//option) any later version.
//This program is distributed in the hope that it will be useful, but
//WITHOUT ANY WARRANTY; without even the implied warranty of
//MERCHANTABILITY or FITNESS FOR A PARTICULAR PURPOSE. See the GNU
//General Public License for more details.
//You should have received a copy of the GNU General Public License along
//with this program. If not, see http://www.gnu.org/licenses/.
header('Content-Type: text/plain; charset=UTF-8');

$outtext = "";
$rules = array(
//Currently 285 rules
'ovnicxki'=>'',
'ovnicxka'=>'',
'ovnika'=>'',
'ovniku'=>'',
'ovnicxe'=>'',
'kujemo'=>'',
'ovacyu'=>'',
'ivacyu'=>'',
'isacyu'=>'',
'dosmo'=>'',
'ujemo'=>'',
'ijemo'=>'',
'ovski'=>'',
'ajucxi'=>'',
'icizma'=>'',
'ovima'=>'',
'ovnik'=>'',
'ognu'=>'',
'inju'=>'',
'enju'=>'',
'cxicyu'=>'',
'sxtva'=>'',
'ivao'=>'',
'ivala'=>'',
'ivalo'=>'',
'skog'=>'',
'ucxit'=>'',
'ujesx'=>'',
'ucyesx'=>'',
'ocyesx'=>'',
'osmo'=>'',
'ovao'=>'',
``` |



```
'ovala'=>'',
'ovali'=>'',
'ismo'=>'',
'ujem'=>'',
'esmo'=>'',
'asmo'=>'',    //pravi gresku kod pevasmo
'zxemo'=>'',
'cyemo'=>'',
'cyemo'=>'',
'bemo'=>'',
'ovan'=>'',
'ivan'=>'',
'isan'=>'',
'uvsxi'=>'',
'ivsxi'=>'',
'evsxi'=>'',
'avsxi'=>'',
'sxucyi'=>'',
'uste'=>'',
'icxe'=>'i',//bilo ik
'acxe'=>'ak',
'uzxe'=>'ug',
'azxe'=>'ag',// mozda treba az, pokazati, pokazxe
'aci'=>'ak',
'oste'=>'',
'aca'=>'',
'enu'=>'',
'enom'=>'',
'enima'=>'',
'eta'=>'',
'etu'=>'',
'etom'=>'',
'adi'=>'',
'alja'=>'',
'nju'=>'nj',
'lju'=>'',
'lja'=>'',
'lji'=>'',
'lje'=>'',
'ljom'=>'',
'ljama'=>'',
'zi'=>'g',
'etima'=>'',
'ac'=>'',
'becyi'=>'beg',
'nem'=>'',
'nesx'=>'',
'ne'=>'',
'nemo'=>'',
'nimo'=>'',
'nite'=>'',
'nete'=>'',
'nu'=>'',
'ce'=>'',
'ci'=>'',
```



```
'cu'=>'',
'ca'=>'',
'cem'=>'',
'cima'=>'',
'sxcyu'=>'s',
'ara'=>'r',
'iste'=>'',
'este'=>'',
'aste'=>'',
'ujte'=>'',
'jete'=>'',
'jemo'=>'',
'jem'=>'',
'jesx'=>'',
'ijte'=>'',
'inje'=>'',
'anje'=>'',
'acxki'=>'',
'anje'=>'',
'inja'=>'',
'cima'=>'',
'alja'=>'',
'etu'=>'',
'nog'=>'',
'omu'=>'',
'emu'=>'',
'uju'=>'',
'iju'=>'',
'sko'=>'',
'eju'=>'',
'ahu'=>'',
'ucyu'=>'',
'icyu'=>'',
'ecyu'=>'',
'acyu'=>'',
'ocu'=>'',
'izi'=>'ig',
'ici'=>'ik',
'tko'=>'d',
'tka'=>'d',
'ast'=>'',
'tit'=>'',
'nusx'=>'',
'cyesx'=>'',
'cxno'=>'',
'cxni'=>'',
'cxna'=>'',
'uto'=>'',
'oro'=>'',
'eno'=>'',
'ano'=>'',
'umo'=>'',
'smo'=>'',
'imo'=>'',
'emo'=>'',
```



```
            'ulo'=>'',
            'sxlo'=>'',
            'slo'=>'',
            'ila'=>'',
            'ilo'=>'',
            'ski'=>'',
            'ska'=>'',
            'elo'=>'',
            'njo'=>'',
            'ovi'=>'',
            'evi'=>'',
            'uti'=>'',
            'iti'=>'',
            'eti'=>'',
            'ati'=>'',
            'vsxi'=>'',
            'vsxi'=>'',
            'ili'=>'',
            'eli'=>'',
            'ali'=>'',
            'uji'=>'',
            'nji'=>'',
            'ucyi'=>'',
            'sxcyi'=>'',
            'ecyi'=>'',
            'ucxi'=>'',
            'oci'=>'',
            'ove'=>'',
            'eve'=>'',
            'ute'=>'',
            'ste'=>'',
            'nte'=>'',
            'kte'=>'',
            'jte'=>'',
            'ite'=>'',
            'ete'=>'',
            'cyi'=>'',
            'usxe'=>'',
            'esxe'=>'',
            'asxe'=>'',
            'une'=>'',
            'ene'=>'',
            'ule'=>'',
            'ile'=>'',
            'ele'=>'',
            'ale'=>'',
            'uke'=>'',
            'tke'=>'',
            'ske'=>'',
            'uje'=>'',
            'tje'=>'',
            'ucye'=>'',
            'sxcye'=>'',
            'icye'=>'',
            'ecye'=>'',
```



```
'ucxe'=>'',
'oce'=>'',
'ova'=>'',
'eva'=>'',
'ava'=>'av',
'uta'=>'',
'ata'=>'',
'ena'=>'',
'ima'=>'',
'ama'=>'',
'ela'=>'',
'ala'=>'',
'aka'=>'',
'aja'=>'',
'jmo'=>'',
//'uga'=>'',
'oga'=>'',
'ega'=>'',
'aća'=>'',
'oca'=>'',
'aba'=>'',
'cxki'=>'',
'ju'=>'',
'hu'=>'',
'cyu'=>'',
'cu'=>'',
'ut'=>'',
'it'=>'',
'et'=>'',
'at'=>'',
'usx'=>'',
'isx'=>'',
'esx'=>'',
'esx'=>'',
'uo'=>'',
'no'=>'',
'mo'=>'',
'mo'=>'',
'lo'=>'',
'ko'=>'',
'io'=>'',
'eo'=>'',
'ao'=>'',
'un'=>'',
'an'=>'',
'om'=>'',
'ni'=>'',
'im'=>'',
'em'=>'',
'uk'=>'',
'uj'=>'',
'oj'=>'',
'li'=>'',
'ci'=>'',
'uh'=>'',
```



```php
    'oh'=>'',
    'ih'=>'',
    'eh'=>'',
    'ah'=>'',
    'og'=>'',
    'eg'=>'',
    'te'=>'',
    'sxe'=>'',
    'le'=>'',
    'ke'=>'',
    'ko'=>'',
    'ka'=>'',
    'ti'=>'',
    'he'=>'',
    'cye'=>'',
    'cxe'=>'',
    'ad'=>'',
    'ecy'=>'',
    'ac'=>'',
    'na'=>'',
    'ma'=>'',
    'ul'=>'',
    'ku'=>'',
    'la'=>'',
    'nj'=>'nj',
    'lj'=>'lj',
    'ha'=>'',
    'a'=>'',
    'e'=>'',
    'u'=>'',
    'sx'=>'',
    'o'=>'',
    'i'=>'',
    //'k'=>'',
    'j'=>'',
    //'t'=>'',
    //'n'=>'', //London, londona
    'i'=>''
);

$dictionary = array(
//biti glagol
    'bih'=>'biti',
    'bi'=>'biti',
    'bismo'=>'biti',
    'biste'=>'biti',
    'bisxe'=>'biti',
    'budem'=>'biti',
    'budesx'=>'biti',
    'bude'=>'biti',
    'budemo'=>'biti',
    'budete'=>'biti',
    'budu'=>'biti',
    'bio'=>'biti',
    'bila'=>'biti',
```



```
'bili'=>'biti',
'bile'=>'biti',
'biti'=>'biti',
'bijah'=>'biti',
'bijasxe'=>'biti',
'bijasmo'=>'biti',
'bijaste'=>'biti',
'bijahu'=>'biti',
'besxe'=>'biti',
//jesam
'sam'=>'jesam',
'si'=>'jesam',
'je'=>'jesam',
'smo'=>'jesam',
'ste'=>'jesam',
'su'=>'jesam',
'jesam'=>'jesam',
'jesi'=>'jesam',
'jeste'=>'jesam',
'jesmo'=>'jesam',
'jeste'=>'jesam',
'jesu'=>'jesam',
// glagol hteti
'cyu'=>'hteti',
'cyesx'=>'hteti',
'cye'=>'hteti',
'cyemo'=>'hteti',
'cyete'=>'hteti',
'hocyu'=>'hteti',
'hocyesx'=>'hteti',
'hocye'=>'hteti',
'hocyemo'=>'hteti',
'hocyete'=>'hteti',
'hocye'=>'hteti',
'hteo'=>'hteti',
'htela'=>'hteti',
'hteli'=>'hteti',
'htelo'=>'hteti',
'htele'=>'hteti',
'htedoh'=>'hteti',
'htede'=>'hteti',
'htede'=>'hteti',
'htedosmo'=>'hteti',
'htedoste'=>'hteti',
'htedosxe'=>'hteti',
'hteh'=>'hteti',
'hteti'=>'hteti',
'htejucyi'=>'hteti',
'htevsxi'=>'hteti',
// glagol moći
'mogu'=>'mocyi',
'možeš'=>'mocyi',
'može'=>'mocyi',
'možemo'=>'mocyi',
'možete'=>'mocyi',
```



```php
'mogao'=>'mocyi',
'mogli'=>'mocyi',
'moći'=>'mocyi'
);

if (isset($_POST["queue"]))
{
    $text= strtolower($_POST["queue"]);
    $text = trim($text);
    $substr1="";
    $substr2="";
    //Tokenizes and set interpucntion marks to be separated with blank char form word
    for($i=0;$i<strlen($text);$i++)
    {
        if(($text[$i]=="." or $text[$i]=="," or $text[$i]=="!" or $text[$i]==":" or $text[$i]=="?" or $text[$i]=="(" or $text[$i]==")" or $text[$i]==";") and $text[$i-1]!=" ")
        {
            $substr1 = substr($text,0,$i);
            $substr2 = substr($text,$i,strlen($text));
            $text = $substr1 . " " . $substr2;
        }
        if(($text[$i]=="." or $text[$i]=="," or $text[$i]=="!" or $text[$i]==":" or $text[$i]=="?" or $text[$i]=="(" or $text[$i]==")" or $text[$i]==";") and $text[$i+1]!=" ")
        {
            $substr1 = substr($text,0,$i+1);
            $substr2 = substr($text,$i+1,strlen($text));
            $text = $substr1 . " " . $substr2;
        }
    }
    //Creates tokens
    $tokens = explode(" ",$text );
    $arrkeys =array_keys($rules);
    //Stemmes
    for($i=0;$i<count($tokens);$i++)
    {
        $currtoken = $tokens[$i];
        //Checks if word is in dictionary, if yes changes to --
        //original form
        if(in_array($currtoken,array_keys($dictionary) )){
            $tokens[$i]=$dictionary[$currtoken];
            $outtext=$outtext . " " . $tokens[$i];
            continue;
        }
        for($j=0;$j<count($rules);$j++)
        {
            if(preg_match('/\b(cx|cy|zx|dx|sx)/',$tokens[$i]))
                $pattern ='/(\w{3,})'.$arrkeys[$j].'\b/';
            else
                $pattern ='/(\w{2,})'.$arrkeys[$j].'\b/';
            $arrkey = $arrkeys[$j];
            //'/(\w{2,})ski\b/'
```



```php
                        if(preg_match($pattern,$tokens[$i]))
                        {
                                $tokens[$i] = preg_replace($pattern,'$1'.$rules[$arrkey] ,$tokens[$i] );
                                break;
                        }
                }
                $outtext=$outtext . " " . $tokens[$i];
        }
}
else
{
        $text= "Error: Not set queue";
        $outtext = $text;
}
//Output
echo $outtext;

?>
```